\documentclass[10pt,twocolumn,letterpaper]{article}

\usepackage[pagenumbers]{iccv}
\definecolor{iccvblue}{rgb}{0.21,0.49,0.74}
\usepackage[pagebackref,breaklinks,colorlinks,allcolors=iccvblue]{hyperref}
\usepackage{multirow}
\usepackage{colortbl}
\usepackage{arydshln}
\usepackage{hhline}
\usepackage{algorithm} 
\usepackage{algpseudocode}

\def\paperID{3712}
\def\confName{ICCV}
\def\confYear{2025}

\title{TeEFusion: Blending Text Embeddings to Distill Classifier-Free Guidance}

\author{Minghao Fu\thanks{Work done during the internship at Alibaba International Digital Commerce Group.}\textsuperscript{\rm ~~1,2,3}~~~Guo-Hua Wang\thanks{G. Wang is the corresponding author.}\textsuperscript{\rm ~~3}~~~Xiaohao Chen\textsuperscript{\rm 3}~~~Qing-Guo Chen\textsuperscript{\rm 3} \\ Zhao Xu\textsuperscript{\rm 3}~~~Weihua Luo\textsuperscript{\rm 3}~~~Kaifu Zhang\textsuperscript{\rm 3} \\
\textsuperscript{\rm 1} School of Artificial Intelligence, Nanjing University \\ \textsuperscript{\rm 2} National Key Laboratory for Novel Software Technology, Nanjing University \hspace{0.3cm} \\ 
\textsuperscript{\rm 3} Alibaba International Digital Commerce Group \\
\small{fumh@lamda.nju.edu.cn, ~~\{wangguohua, xiaohao.cxh, qingguo.cqg, changgong.xz, weihua.luowh, kaifu.zkf\}@alibaba-inc.com}
}

\if 0
\author{First Author\\
Institution1\\
Institution1 address\\
{\tt\small firstauthor@i1.org}
\and
Second Author\\
Institution2\\
First line of institution2 address\\
{\tt\small secondauthor@i2.org}
}
\fi

\begin{document}
\maketitle

\begin{abstract}
Recent advances in text-to-image synthesis largely benefit from sophisticated sampling strategies and classifier-free guidance (CFG) to ensure high-quality generation. However, CFG's reliance on two forward passes, especially when combined with intricate sampling algorithms, results in prohibitively high inference costs. To address this, we introduce TeEFusion (\textbf{Te}xt \textbf{E}mbeddings \textbf{Fusion}), a novel and efficient distillation method that directly incorporates the guidance magnitude into the text embeddings and distills the teacher model's complex sampling strategy. By simply fusing conditional and unconditional text embeddings using linear operations, TeEFusion reconstructs the desired guidance without adding extra parameters, simultaneously enabling the student model to learn from the teacher's output produced via its sophisticated sampling approach. Extensive experiments on state-of-the-art models such as SD3 demonstrate that our method allows the student to closely mimic the teacher's performance with a far simpler and more efficient sampling strategy. Consequently, the student model achieves inference speeds up to 6$\times$ faster than the teacher model, while maintaining image quality at levels comparable to those obtained through the teacher's complex sampling approach. The code is publicly available at \href{https://github.com/AIDC-AI/TeEFusion}{github.com/AIDC-AI/TeEFusion}.
\end{abstract}

\section{Introduction}

Models based on the Flow Matching~\cite{rectified_flow,flow_matching} pipeline, such as SD3~\cite{sd3} and FLUX~\cite{flux}, have emerged as leading generative frameworks by directly synthesizing structured images from textual prompts. In the realm of text-to-image generation, their versatility is evidenced by a broad spectrum of applications, ranging from image editing~\cite{instructpix2pix} to creative concept generation for digital art~\cite{dalle2}.

\begin{figure}
    \centering
    \begin{subfigure}[b]{0.4\linewidth}
        \centering
        \includegraphics[width=\linewidth]{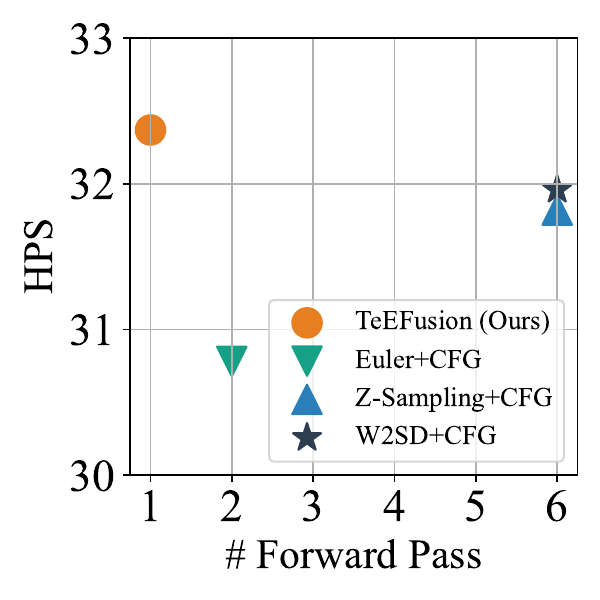}
        \caption{SD3~\cite{sd3}}
        \label{fig:insight_sd3}
    \end{subfigure}
    \hfill
    \begin{subfigure}[b]{0.4\linewidth}
        \centering
        \includegraphics[width=\linewidth]{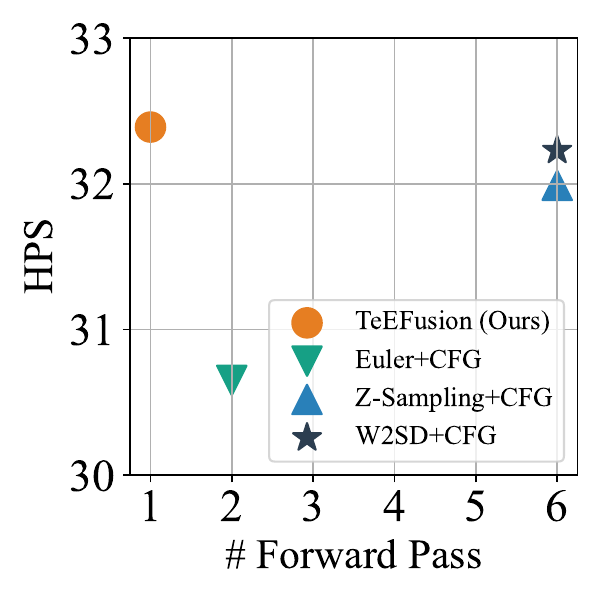}
        \caption{In-house T2I}
        \label{fig:insight_in_house_t2i}
    \end{subfigure}
    \caption{Comparison of different sampling strategy on two DiT~\cite{dit} models. ``\# Forward Pass'' specifies the number of forward passes required in one denoising step.}
    \label{fig:insight}
\end{figure}

\begin{figure*}[t]
    \centering
    \includegraphics[width=.72\linewidth]{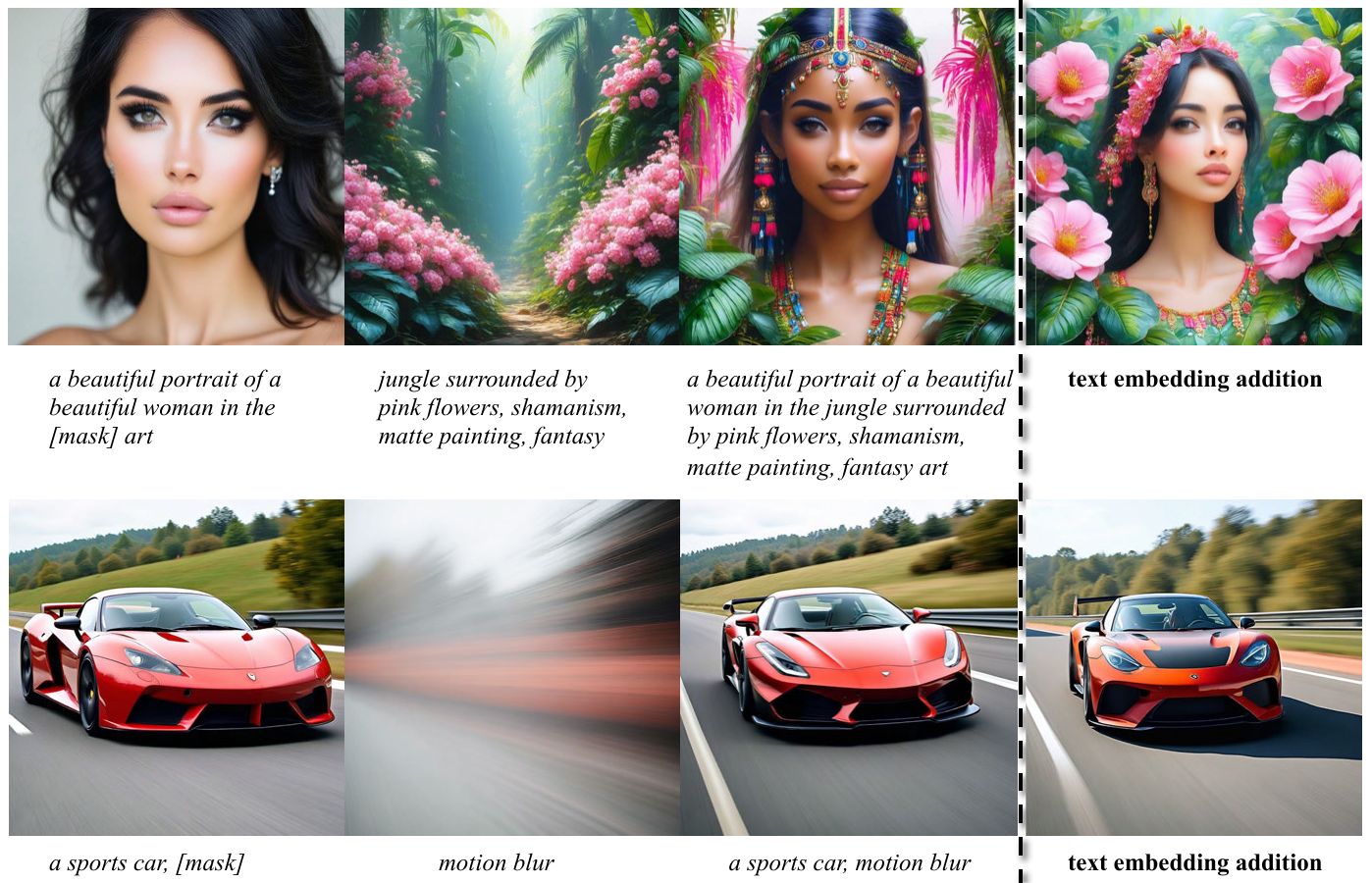}
    \caption{Qualitative results by different prompting strategies. The leftmost three columns display images generated from different prompt inputs, while the final column illustrates the outcome of additively fusing semantic information from both the preserved and masked components.}
    \label{fig:in_depth_analysis}
\end{figure*}

Classifier-Free Guidance (CFG)~\cite{cfg} is a key technique for ensuring high-quality image synthesis by utilizing conditional and unconditional predictions to steer the generation toward regions with higher text-conditioned density. However, sampling with CFG demands two forward passes, one conditioned on the provided textual prompt and another on a null (or background) prompt, which introduces substantial computational overhead and slows down inference.

To mitigate this issue, guidance distillation~\cite{efficient_diffusion_survey} is commonly employed in state-of-the-art models, such as variants like FLUX.1-dev~\cite{flux}. This approach integrates the guidance as an input parameter, reducing the number of required forward passes back to one. Consequently, guidance distillation has become both popular and crucial in state-of-the-art models, as it allows models to be scaled up without concern for cost or to employ more complex inference sampling algorithms. As illustrated in Fig.~\ref{fig:insight}, advanced sampling strategies such as Z-Sampling~\cite{z_sampling}+CFG and W2SD~\cite{w2sd}+CFG require almost 3$\times$ inference burden of the traditional Euler~\cite{rectified_flow}+CFG counterpart, even though they yield higher image quality (as evidenced by HPS~\cite{hpsv2}), thereby highlighting the significant potential for efficiency optimization.

Several recent studies~\cite{dmd,f_divergence,gdd,sid,snapfusion,distill_cfg,progressive_distill} attempt to advance distillation for image synthesis~\cite{efficient_diffusion_survey}. However, these approaches are confined to scenarios where both teacher and student models utilize the same sampling algorithm. In other words, they do not focus on distilling from particularly complex sampling algorithms while enabling the student model to rely solely on a simple sampling strategy. Moreover, these methods are overly complex, posing significant obstacles for implementation and adaptation. This issue is further amplified by the large-scale nature of modern state-of-the-art models, where tuning hyperparameters becomes increasingly challenging, ultimately hindering the deployment of these intricate methods.

As a result, these methods currently lack robust and large-scale validation, particularly in two critical aspects: 1) the inclusion of general benchmarks, such as DPG-Bench~\cite{dpg_bench} or aesthetic scoring metrics like HPS~\cite{hpsv2}; and 2) testing on architectures that match the scale of the large DiT~\cite{dit} used in state-of-the-art models like SD3~\cite{sd3} and FLUX~\cite{flux}. Consequently, it remains an open question whether these distillation techniques can be effectively applied to the latest, most advanced text-to-image generation models.

Motivated by these challenges, in this paper we explore more effective and efficient approaches for CFG distillation. We find that linear operations on different text features in the embedding space can effectively fuse or suppress aspects of the original text (cf. Fig.~\ref{fig:in_depth_analysis}). Based on this observation, we propose TeEFusion (\textbf{Te}xt \textbf{E}mbeddings \textbf{Fusion}), a novel yet remarkably simple distillation method that directly leverages the underlying sampling formulation of CFG and integrates the guidance magnitude into the text embeddings. Specifically, our approach moves the linear combination of predictions from the conditional and unconditional model outputs forward, applying it directly to combine the corresponding text embeddings. The student model is optimized to learn the teacher model's denoised outputs, generated via a complex sampling strategy, thereby concurrently distilling both guidance signals and sampling procedures.

\begin{figure*}[t]
    \centering
    \includegraphics[width=.85\linewidth]{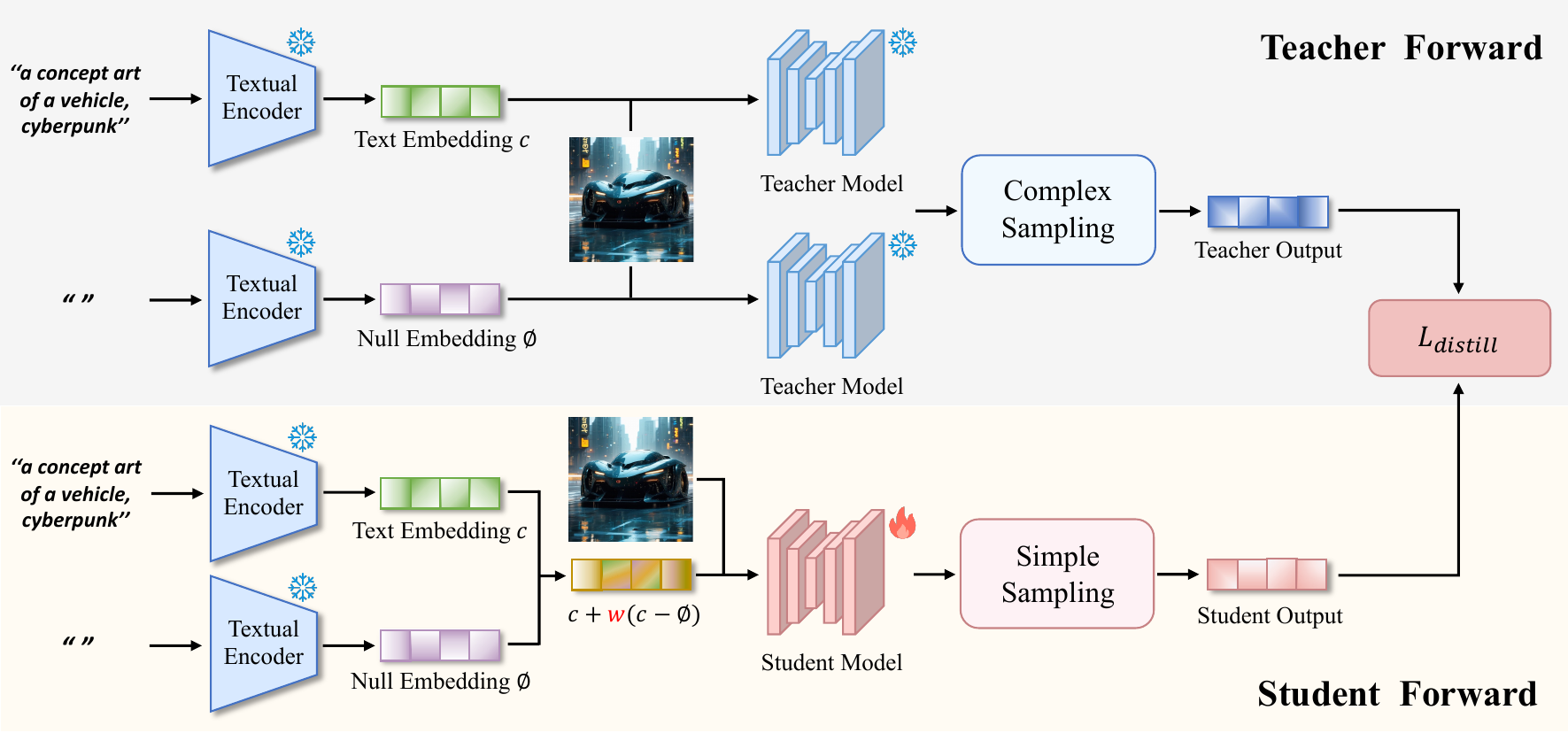}
    \caption{Illustration of TeEFusion. Our approach shifts the fusion of the DiT~\cite{dit} model's outputs directly into the text embedding space, as specified in Eq.~\ref{eq:teefusion}. This design is compatible with various sampling methods employed by the teacher model, offering significant potential for test-time scaling.}
    \label{fig:network}
\end{figure*}

TeEFusion offers two primary advantages. First, compared to existing distillation techniques, our approach is exceptionally simple, with clear and easy-to-implement algorithmic principles. Unlike~\cite{distill_cfg}, which employs a feed-forward network to bridge the guidance magnitude, our method introduces no additional architecture—the structure of the distilled model remains identical to the undistilled model. Moreover, TeEFusion does not introduce any extra hyperparameters, ensuring that the distillation process remains efficient and easy to integrate into existing pipelines.

Secondly, we conduct a comprehensive evaluation of TeEFusion on state-of-the-art text-to-image models, including SD3~\cite{sd3} and our In-house T2I, which demonstrates performance on par with SD3. Empirically results demonstrate that TeEFusion achieves significant improvements over baseline methods in terms of aesthetic scoring~\cite{hpsv2}, object composition~\cite{dpg_bench} and prompt-following ability~\cite{clip}. Moreover, our approach seamlessly accommodates the diverse samplers used by the teacher model. We hope that our evaluations will provide valuable insights for assessing distillation techniques in state-of-the-art, large-scale text-to-image models.

The contributions of this paper can be summarized as:
\begin{itemize}
    \item We empirically find that for current state-of-the-art text-to-image generation models, adding or subtracting text embeddings from specific prompts effectively merges or alleviates certain visual concepts in generated images.

    \item We propose TeEFusion, a novel and simple distillation strategy that incorporates guidance from CFG directly into the text embeddings without introducing any extra parameters. Our approach is seamlessly compatible with a variety of complex sampling strategies employed by the teacher model.

    \item We evaluate TeEFusion on industrial-grade, large-scale text-to-image models, including state-of-the-art frameworks such as SD3~\cite{sd3} and our In-house T2I model, using standard benchmarks like DPG-Bench~\cite{dpg_bench}, aesthetic metrics (HPS~\cite{hpsv2}), and prompt-following assessments~\cite{clip}. Empirical results demonstrate that TeEFusion significantly outperforms baseline methods in terms of aesthetic quality, object composition, and prompt adherence.
\end{itemize}

\section{Related Work}

\subsection{Test-Time Scaling}

Recent research has increasingly focused on the scaling behavior of inference in diffusion models~\citep{cfg,repaint}. RePaint~\citep{repaint} iteratively refined latent variables by reintroducing Gaussian noise to restore previous noise levels, establishing a reflection-based sampling paradigm that has been widely adopted in subsequent works~\cite{universial_guidacne,tav,z_sampling,w2sd}. For example, Z-Sampling~\cite{z_sampling} changed denoised outputs in a zigzag manner by alternating forward and reverse passes with varying guidance magnitudes, thereby steering the random input toward semantically meaningful regions. W2SD~\cite{w2sd} further extended this approach by explicitly incorporating both strong and weak models. 

\subsection{Diffusion Distillation}

Early studies aimed to enhance the sampling efficiency of diffusion models by leveraging advanced numerical solvers~\cite{ddim,pseudo,dpm}. Despite these improvements, significant room for further efficiency gains remains. Diffusion distillation~\cite{efficient_diffusion_survey} addresses this challenge by training a simpler student model to replicate the behavior of a more complex teacher model, primarily through step distillation~\cite{add,ladd,progressive_distill,edm,operator_learning,consistency_model}, thereby reducing the required inference steps. Some approaches~\cite{add,ladd} incorporated adversarial training by introducing an additional discriminator to facilitate generalization of the score function. However, the complexity inherent to these methods limits their practical deployment, especially given the scale and complexity of contemporary diffusion models. 

Among these methods, DistillCFG~\cite{distill_cfg} is the most relevant, using an extra MLP to encode the guidance scale. DICE~\cite{dice} also adds an attention-enhanced MLP but distills only a single fixed scale, which limits flexibility at inference. More variants such as MG~\cite{tang2025diffusion} remain unevaluated on industrial-scale datasets and models. Additionally, existing distillation frameworks, such as progressive distillation~\cite{progressive_distill}, have not explicitly addressed scenarios involving teacher models that adopt complex test‐time sampling strategies, a situation increasingly prevalent in state‐of‐the‐art diffusion models.

To bridge this critical gap, our proposed framework, TeEFusion, advances guidance distillation by explicitly incorporating the guidance magnitude into the linear combination of conditional and unconditional text embeddings. Furthermore, TeEFusion effectively generalizes across diverse complex sampling strategies employed by teacher models, addressing an essential yet largely overlooked aspect of diffusion model distillation.

\section{Preliminaries}

\subsection{Classifier-Free Guidance}
Classifier-Free Guidance (CFG)~\cite{cfg} enhances the alignment between generated images and textual descriptions by adjusting the sampling distribution:
\begin{equation}
    \label{eq:cfg_sampling}
    \tilde{p}_{\theta}(x_t|c) \propto p_{\theta}(x_t|c)^{1+w} p_{\theta}(x_t)^{-w},
\end{equation}
where $c$ represents the text prompt, $p_{\theta}(x_t|c)$ is the conditional distribution, and $p_{\theta}(x_t)$ denotes the unconditional counterpart. CFG refines the generation process by estimating the diffusion score\footnote{For notational simplicity, we denote the score prediction for both diffusion and flow matching models as $\epsilon_\theta$.} as:
\begin{equation}
\label{eq:cfg_score}
\tilde{\epsilon}_{\theta}(x_t, c) = (1+w)\epsilon_{\theta}(x_t, c) - w\epsilon_{\theta}(x_t),
\end{equation}

The scalar $w$ determines the strength of guidance: larger values steer the synthesis more strongly towards the textual prompt, thereby promoting outputs that faithfully reflect the input text while reducing dependence on unconditioned predictions. Note that two forward passes are required to get the prediction from $\epsilon_{\theta}(x_t, c)$ and $\epsilon_{\theta}(x_t)$.

\subsection{Sampling with Reflection.}
Some methods~\cite{z_sampling,w2sd} employ sampling strategies that extend beyond the standard Euler method with reflection, refining the starting points of the denoising process with richer semantic information.

Weak-to-Strong Diffusion (W2SD)~\cite{w2sd} leverages the discrepancy between a weak model $\mathcal M^w$ and a strong model $\mathcal M^s$ to bridge the gap toward an ideal generative distribution. In W2SD, a reflective operation updates the latent variable $x_t$ as follows:
\begin{align}
    \label{eq:go}
    \tilde x_t &= \mathcal M_{\text{inv}}^w (\mathcal M^s(x_t, t), t) \\
    \label{eq:back}
    x_{t-1} &= \mathcal M^s (\tilde x_t, t ),
\end{align}
thereby guiding the sampling trajectory toward regions of higher text density. Z-Sampling~\cite{z_sampling} is a simpler version of W2SD by sharing the weights between $\mathcal M^s$ and $\mathcal M^k$, but using different guidance signals.

However, the extra inversion step where the weak model processes the denoised output of the strong models 
introduces two additional forward passes per sampling iteration. Consequently, the time cost of reflection sampling is 3$\times$ compared to standard sampling (one for standard denoising, one for inversion, and one for subsequent denoising). When combined with CFG, the overall time cost escalates to nearly 6$\times$.

\begin{algorithm}[t]
\centering
\caption{The distillation pipeline of TeEFusion.}\label{alg:stage1}
\begin{algorithmic}
\Require Teacher model $\epsilon_{\theta_{\mathrm{T}}}$, Dataset $D$
\State $\epsilon_{\theta_{\mathrm S}} \gets \epsilon_{\theta_{\mathrm T}}$ \Comment{Initialize student using teacher weights}
\While{not converged}
\State $x_0, c \sim \mathcal{D}$ \Comment{Sample data}
\State $t \sim U[0, 1]$ \Comment{Sample time}
\State $w \sim U[w_{\text{min}}, w_{\text{max}}]$ \Comment{Sample guidance}
\State $\epsilon \sim N(0, I)$ \Comment{Sample noise}
\State $x_t = (1-t)x_0 + t\epsilon$ \Comment{Add noise to data}
\State $\tilde \epsilon_{\theta_{\mathrm{T}}} (x_t,w,c) \!=\! (1+w)\epsilon_{\theta_{\mathrm{T}}}(x_t, t, c) \!-\! w \epsilon_{\theta_{\mathrm{T}}}(x_t, t, \varnothing)$ \Comment{Compute target using CFG}
\If{Reflection}
    \State $\tilde \epsilon_{\theta_{\mathrm{T}}}(x_t,w,c) \gets$  Refine $\tilde \epsilon_{\theta_{\mathrm{T}}}(x_t,w,c)$ using Eq.~\ref{eq:go} and~\ref{eq:back}
\EndIf
\State Compute $\hat z_{t,c,\varnothing,w}$ using Eq.~\ref{eq:teefusion}
\State $L_{\text{distll}} = \lVert \epsilon_{\theta_\mathrm{S}} (x_t, \hat z_{t,c,\varnothing,w}) - \tilde \epsilon_{\theta_{\mathrm{T}}}(x_t,w,c) \rVert_{2}^{2}$ \Comment{Loss}
\State $\epsilon_{\theta_\mathrm{S}} \gets \epsilon_{\theta_\mathrm{S}} - \gamma \cdot \nabla_{\epsilon_{\theta_\mathrm{S}}} L_{\text{distill}}$ \Comment{Optimization}
\EndWhile
\State \Return $\epsilon_{\theta_{\mathrm{S}}}$ \Comment{Output the trained student model}
\end{algorithmic}
\end{algorithm}

\section{An In-Depth Analysis of CFG}
\label{sec:analysis}

We begin by examining the mechanism underlying CFG. As indicated by Eq.~\ref{eq:cfg_score}, the output of CFG can be expressed as a linear combination of the conditional prediction $\epsilon_\theta(x_t|c)$ (with coefficient $1+w$) and the unconditional prediction $\epsilon_\theta(x_t)$ (with coefficient $-w$). This formulation motivates the idea of moving this linear combination earlier in the processing pipeline, thereby reducing the need for two separate forward passes to just one. A natural approach is to integrate this combination directly into the text embeddings to efficiently incorporate the guidance magnitude into the model as:
\begin{equation}
\label{eq:prompt_cfg}
    \hat{\epsilon}_{\theta}(x_t, c) = \epsilon_{\theta}(x_t, \hat c),
\end{equation}
where $\hat c = (1+w)c - w\varnothing=c+w(c-\varnothing)$ is marked as the \textit{fusion text embedding} with $\varnothing$ from the null prompt to provide the background for generation. However, the success of this approach depends on the assumption that linearly combining input conditions is meaningful. 

Therefore, we perform a preliminary test to verify whether merging the conditional and unconditional text embeddings into a single text embedding can yield the necessary semantic representations. In this experiment, we randomly masked certain components of the prompt, replacing them with a \textit{[mask]} token. We then generated images under three different conditions: 
\begin{enumerate}
    \item Using only the unmasked portion of the prompt.
    \item Using only the masked components (i.e., the content that was replaced by \textit{[mask]}).
    \item Using the original, unaltered prompt.
    \item Both the unmasked and masked components are converted into text embeddings, which are then merged via element-wise addition. The resulting fused embedding is subsequently used for image generation.
\end{enumerate}

This setup enables us to evaluate whether the additive (or subtractive) fusion of semantic information from both preserved and masked prompt components yields meaningful features that can guide high-quality image synthesis via a linear combination (cf. Eq.~\ref{eq:prompt_cfg}) in the text embedding space. As shown in Fig.~\ref{fig:in_depth_analysis}, images generated by adding the text embeddings of the preserved and masked parts match (or even surpass) the quality of those produced from the original, unaltered prompt. These findings indicate that additive operations in the text embedding space can effectively merge diverse prompt patterns, thereby highlighting the potential of our sampling technique to use Eq.~\ref{eq:prompt_cfg} as a surrogate to integrate guidance magnitudes while keeping high efficiency during sampling (see Sec.~\ref{sec:cosine_similarity} for quantitative results).

\section{TeEFusion to Distill CFG}

The analysis in Sec.~\ref{sec:analysis} demonstrates that simple addition or subtraction of text embeddings from different prompts can effectively reconstruct or eliminate specific semantic components of the original prompt, indicating that such operations in the embedding space yield features with clear semantics. Building on this observation, we hypothesize that any linear combination of these text embeddings—with appropriately moderate coefficients can similarly produce semantically meaningful representations. This insight motivates the use of Eq.~\ref{eq:prompt_cfg} for guidance distillation with a properly configured $w$.

Motivated by this hypothesis, we propose TeEFusion (\textbf{Te}xt \textbf{E}mbeddings \textbf{Fusion}), a simple distillation framework that effectively distills both classifier-free guidance and complex reflection-based sampling techniques simultaneously. In TeEFusion, guidance magnitude $w$ is incorporated into the student model by linearly combining conditional and unconditional text embeddings, with $w$ serving as the combination coefficient. Subsequently, the student model is trained to mimic the denoised output of teacher models that employ reflection-based sampling (see Fig.~\ref{fig:network}).

However, directly applying Eq.~\ref{eq:prompt_cfg} to TeEFusion poses challenges. As the guidance scale $w$ increases, the variance of the fused prompt $\hat{c}$ grows on the order of $\mathcal{O}(w^2)$, which, when $w$ becomes too large, leads to poor numerical stability. To address this issue, we project $w$ into a vector space using a sine and cosine time embedding~\cite{transformer}. This approach not only ensures that different values of $w$ remain distinguishable, but also mitigates the numerical instability arising from excessive variance.

Specifically, the joint embedding of the text prompt and timestep in the text-to-image model~\cite{sd3,flux} $\epsilon(x_t, t, c)$ is formulated as:
\begin{equation}
z_{t,c} = \mathcal{G}(\psi(t)) + \mathcal{F}(c),
\end{equation}
where $\mathcal{F}(\cdot)$ and $\mathcal{G}(\cdot)$ denote two distinct multi-layer perceptions (MLPs), and $\psi(\cdot)$ represents the sinusoidal encoding of the timestep. This formulation effectively decouples the processing of textual and temporal information, thereby enabling a more structured integration of the guidance magnitude. 

Our TeEFusion distills the guidance magnitude $w$ by incorporating it into $z_{t,c}$ as:
\begin{equation}
\label{eq:teefusion}
\hat z_{t,c,\varnothing,w} = \mathcal{G}(\psi(t)) + \mathcal{F}(c) + \underbrace{\mathcal{G}(\psi(w))\, \mathcal{F}(c - \varnothing)}_{\text{extra term}},
\end{equation}
where the term $c-\varnothing$ is computed after projection into the text embedding space. Assume that the teacher and student models are denoted by $\epsilon_{\theta_{\mathrm{T}}}$ and $\epsilon_{\theta_{\mathrm{S}}}$, respectively. After obtaining $\hat z_{t,c,\varnothing,w}$, it is used to compute the guided output of student model. The distillation objective then lets the student mimic the CFG-derived output from the teacher model with complex sampling (cf. Algorithm~\ref{alg:stage1}).
 
Comparing Eq.~\ref{eq:teefusion} with the definition of $\hat c$, the term $\mathcal{F}(c) +\mathcal{G}(\psi(w))\mathcal{F}(c - \varnothing)$ exactly corresponds to the term $c + w(c-\varnothing)$ in $\hat c$, but computed in the embedding space. 

TeEFusion offers several notable advantages. First, its $w$-injection method is exceptionally simple. Unlike other approaches~\cite{distill_cfg,add,ladd} that require extra network structures and parameters, our method is very easy to implement. Second, it demonstrates strong scalability. By employing more robust teacher sampling strategies or larger network architectures, our approach can further empower the student model. Finally, it is easy to train and converges quickly, as evidenced by the results in Fig.~\ref{fig:ablate_loss_sub}. These features make TeEFusion an attractive and practical solution for efficient guidance distillation in state-of-the-art text-to-image generation systems.

\begin{figure*}[t]
    \centering
    \includegraphics[width=.8\linewidth]{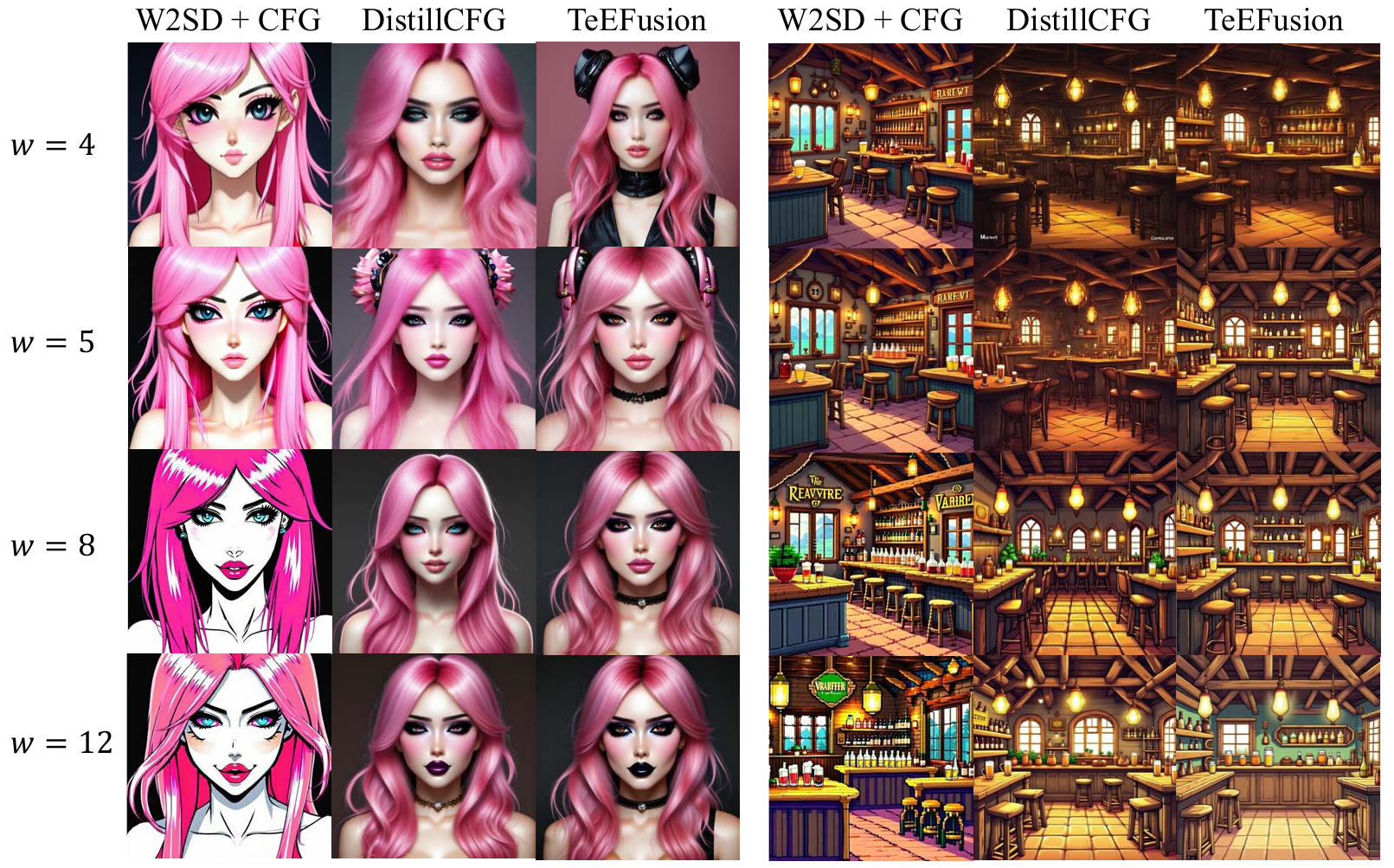}
    \caption{Qualitative comparison among various methods across different guidance scales $w$. Prompts: 1) \textit{An egirl with pink hair and extensive makeup.} 2) \textit{A cozy tavern with a retro video game vibe and cinematic lighting, featuring a cartoon-style animation and detailed background art.}}
    \label{fig:w_compare}
\end{figure*}

\section{Experiments}

\begin{table}[t]
 \centering
 \small
 \caption{HPS results ($\uparrow$) on aesthetic benchmarks. Higher scores indicate that the generated images better conform to human aesthetic standards. ``Cost'' refers to the inference time required, measured and compared separately for each model. ``I.-T2I'' is the abbreviation for ``In-house T2I''. Results of the student model are from the teacher model using W2SD+CFG. The best results are in bold face.}
 \setlength{\tabcolsep}{1.5pt}
 \begin{tabular}{ll|c|cccc}
 \hline
 \multirow{2}{*}{Model} & \multirow{2}{*}{Method} & \multirow{2}{*}{Cost} & \multicolumn{4}{c}{Prompt Group} \\
  & & & Anime & Concept-Art & Paintings & Photo \\
 \hline
 \multirow{6}{*}{SD3} &\multicolumn{6}{l}{\cellcolor{gray!15}\emph{Teacher}}\\
 \cline{2-7}
    & CFG                                   & 2$\times$ & 30.78  & 30.06 & 30.28 & 27.93  \\
    & W2SD+CFG                              & 6$\times$ & 31.96  & 30.65 & 30.67 & 29.76  \\
    \hhline{~------}
    & \multicolumn{6}{l}{\cellcolor{gray!15} \emph{Student}}\\
    \cline{2-7}
    & DistillCFG                            & 1$\times$ & 31.14  & 29.52 & 30.03 & 29.04  \\
    & TeEFusion                             & 1$\times$ & \textbf{32.37}  & \textbf{30.88} & \textbf{30.74} & \textbf{29.84}  \\
    \hline
    \multirow{6}{*}{I.-T2I}  &\multicolumn{6}{l}{\cellcolor{gray!15}\emph{Teacher}}\\
    \cline{2-7}
    & CFG                                   & 2$\times$ & 30.65  & 29.27  & 28.94  & 27.99 \\
    & W2SD+CFG                              & 6$\times$ & 32.23  & 31.29  & 31.22  & 29.93 \\
    \hhline{~------}
    & \multicolumn{6}{l}{\cellcolor{gray!15}\emph{Student}}\\
    \cline{2-7}
    & DistillCFG                            & 1$\times$ & 30.80  & 29.41  & 29.14  & 28.79 \\
    & TeEFusion                             & 1$\times$ & \textbf{32.39}  & \textbf{31.34}  & \textbf{31.27} & \textbf{29.96} \\
 \hline
 \end{tabular}
 \label{tab:aesthetic}
\end{table}

\begin{table*}[htbp!]
\centering
\small
\caption{Quantitative results on CLIP score and DPG-Bench (\%). Higher scores indicate that the generated images exhibit superior overall performance, reflecting enhanced visual quality, finer details, and better alignment with the input prompt. $\dagger$: measured using prompt from Anime.}
\label{tab:dpgbench}
\begin{tabular}{ll|c|ccccc|c}
\hline
\multirow{2}{*}{Model} & \multirow{2}{*}{Method} & \multirow{2}{*}{CLIP$^\dagger$} & \multicolumn{6}{c}{DPG-Bench} \\
\cline{4-9}
& & & Global $\uparrow$ & Entity $\uparrow$ & Attribute $\uparrow$ & Relation $\uparrow$ & Other $\uparrow$ & Overall $\uparrow$ \\
\hline
\multirow{6}{*}{SD3} &\multicolumn{8}{l}{\cellcolor{gray!15}\emph{Teacher}}\\
    \cline{2-9}
    & CFG                    & 34.93 & 85.71 & 90.84 & 87.79 & 93.58 & 86.40 & 85.09 \\
    & W2SD+CFG               & 34.96 & 82.98 & 92.13 & 88.60 & 94.24 & 91.60 & 86.56 \\
    \hhline{~--------}
    &\multicolumn{8}{l}{\cellcolor{gray!15}\emph{Student}}\\
    \cline{2-9}
    & DistillCFG             & 34.53 & 81.46 & 90.47 & 87.85 & \textbf{93.42} & 84.40 & 84.13 \\
    & TeEFusion              & \textbf{34.63} & \textbf{81.76} & \textbf{90.60} & \textbf{88.11} & 93.15 & \textbf{86.80} & \textbf{84.56} \\
\hline
\multirow{6}{*}{In-house T2I}  &\multicolumn{8}{l}{\cellcolor{gray!15}\emph{Teacher}}\\
    \cline{2-9}
    & CFG                         & 34.25 & 83.89  & 88.08 & 88.07 & 91.57 & 76.40 & 81.88 \\
    & W2SD+CFG         & 34.62 & 82.37   & 91.41 & 89.08 & 93.73 & 78.80 & 85.20 \\ 
    \hhline{~--------}
    &\multicolumn{8}{l}{\cellcolor{gray!15}\emph{Student}}\\
    \cline{2-9}
    & DistillCFG          & 33.38 & \textbf{84.50}  & \textbf{90.21} & 88.24 & 93.54 & 80.00 & 83.52 \\
    & TeEFusion                              & \textbf{33.81} & 84.19  & 90.08 & \textbf{88.56} & \textbf{93.73} & \textbf{81.60} & \textbf{84.13} \\
\hline
\end{tabular}
\end{table*}

In this section, we begin by describing the experimental setups. Next, we present the main results, demonstrating the effectiveness of TeEFusion through both quantitative and qualitative analyses. Finally, we conduct comprehensive ablation studies to investigate the contributions of each component in our approach and to evaluate its robustness under various configurations.

\subsection{Experimental Settings}

\textbf{Models.} We employ two text-to-image generation models that both leverage the DiT~\cite{dit} architecture and Flow Matching~\cite{rectified_flow} framework: SD3~\cite{sd3} and our In-house T2I. SD3~\cite{sd3} is a publicly available open source model with 2B parameters, offering superior visual quality and enhanced compositional consistency. In-house T2I, is optimized for photorealistic images in e-commerce scenarios and consists of 1B parameters.

\textbf{Training Details.} We employ the subset of LAION-5B~\cite{laion} for distillation. In particular, high-resolution images with an aesthetic score~\cite{laion} exceeding 6 are selected as the training set. The student model is optimized using AdamW~\cite{adamw} with an effective batch size of 128 and a default learning rate of $5 \times 10^{-6}$. For the teacher model, we adopt W2SD~\cite{w2sd}+CFG~\cite{cfg} as the sampling strategy, given its superior performance as demonstrated in Fig.~\ref{fig:insight}. Since W2SD leverages two models of differing strengths during sampling, these models are obtained using CHATS~\cite{chats}. All distilled models in our experiments are trained within the W2SD+CFG framework, with $w_{\text{min}}$ and $w_{\text{max}}$ in Algorithm~\ref{alg:stage1} set as 2 and 14, respectively.

\textbf{Baselines.} Given the limited research on directly applying guidance distillation to large text-to-image models, we adopt DistillCFG~\cite{distill_cfg} as our baseline. DistillCFG integrates the guidance scale through an additional MLP embedding as in recent FLUX.1-dev~\cite{flux}. Moreover, we compare TeEFusion with the teacher model to further validate its effectiveness.

\textbf{Evaluation Details.} \textit{HPS}~\cite{hpsv2} provides a broad benchmark for aesthetic evaluation, featuring 3,200 prompts distributed equally among four distinct styles: \textit{Anime}, \textit{Concept-Art}, \textit{Paintings} and \textit{Photo} (800 prompts each). This extensive prompt collection supports robust and consistent evaluation outcomes. \textit{DPG-Bench}~\cite{dpg_bench} includes 1,065 prompts, with each prompt describes several objects with different features and explains how these objects relate to each other. It is designed to test how well text-to-image models can combine these elements into a coherent image. For aesthetic evaluation, we employ three state-of-the-art evaluators: \textit{HPS}~\cite{hpsv2} (with its v2 version), \textit{ImageReward (IR)}~\cite{image_reward}, and \textit{PickScore}~\cite{pickscore}. We also use the \textit{CLIP}~\cite{clip} score to assess how well the generated images align with their corresponding input prompts.

\begin{table}
    \centering
    \small
    \setlength{\tabcolsep}{4pt}
    \caption{Modular ablation of TeEFusion using In-house T2I with prompts from Anime. The ``Config'' of TeEFusion specifies the configuration of ``\textit{extra term}'' in Eq.~\ref{eq:teefusion}.}
    \begin{tabular}{l|cccc}
    \hline
    \multirow{2}{*}{Config} & \multicolumn{4}{c}{Metrics} \\
    & HPS $\uparrow$ & IR $\uparrow$ & PickScore $\uparrow$ & CLIP $\uparrow$ \\
    \hline
    DistillCFG & 30.80 & 113.12 & 22.42 & 33.38 \\
    \hdashline
    $\mathcal{G}(\psi(w))$ & 31.05 & 116.39 & 22.49 & 33.59  \\
    $\mathcal{G}(\psi(w))\, \mathcal{F}(c - \varnothing)$ & 32.39 & 128.50 & 22.68 & 33.81 \\
    \hline
    \end{tabular}
    \label{tab:modular_ablation}
\end{table}

\subsection{Main Results.}

As demonstrated in Table~\ref{tab:aesthetic}, our TeEFusion method consistently achieves the best scores on aesthetic evaluations, surpassing the DistillCFG baseline and even the teacher model across all prompt categories for both SD3 and In-house T2I models. Table~\ref{tab:dpgbench} further highlights TeEFusion's superior performance on the DPG-Bench, where it not only excels in object attributes, spatial relationships, and overall composition, but also achieves higher CLIP scores compared to DistillCFG, underscoring its ability to faithfully capture textual semantics.

These results clearly indicate that TeEFusion effectively integrates the guidance magnitude through linear fusion in the text embedding space, thereby preserving semantic structure and compositional quality while significantly improving sampling efficiency. Notably, TeEFusion achieves an inference speed that is 6$\times$ faster than that of the teacher model. Overall, TeEFusion successfully distills complex sampling strategies into a streamlined inference process, delivering high image quality with minimal trade-offs, and making it a promising approach for real-world applications.

\begin{table}
    \centering
    \small
    \setlength{\tabcolsep}{3pt}
    \caption{Ablation study on distilling with different sampling strategies. In each group, the first row shows the teacher's sampling strategy, while the second row displays TeEFusion distilled with it. Model: In-house T2I. Prompt: Anime.}

    \begin{tabular}{l|c|cccc}
    \hline
    \multirow{2}{*}{Sampling} & \multirow{2}{*}{Cost} & \multicolumn{4}{c}{Metrics} \\
    & &  HPS $\uparrow$ & IR $\uparrow$ & PickScore $\uparrow$ & CLIP $\uparrow$ \\
    \hline
    Euler+CFG  & 2$\times$ & 30.65 & 118.05 & 22.79 & 34.25 \\
    TeEFusion  & 1$\times$ & 30.61 & 117.29 & 22.78 & 33.58 \\
    \hline
    Z-Sampling+CFG & 6$\times$ & 31.99 & 125.42 & 22.61 & 34.47 \\
    TeEFusion      & 1$\times$ & 32.01 & 125.07 & 22.65 & 33.62 \\
    \hline
    W2SD+CFG       & 6$\times$ & 32.23 & 127.56 & 22.49 & 34.62 \\
    TeEFusion      & 1$\times$ & 32.39 & 128.50 & 22.68 & 33.81 \\
    \hline
    \end{tabular}
    \label{tab:sampling_strategy_ablation}
\end{table}

\subsection{Ablation Studies}

\begin{figure}
    \centering
    \begin{subfigure}[b]{0.41\linewidth}
        \centering
        \includegraphics[width=\linewidth]{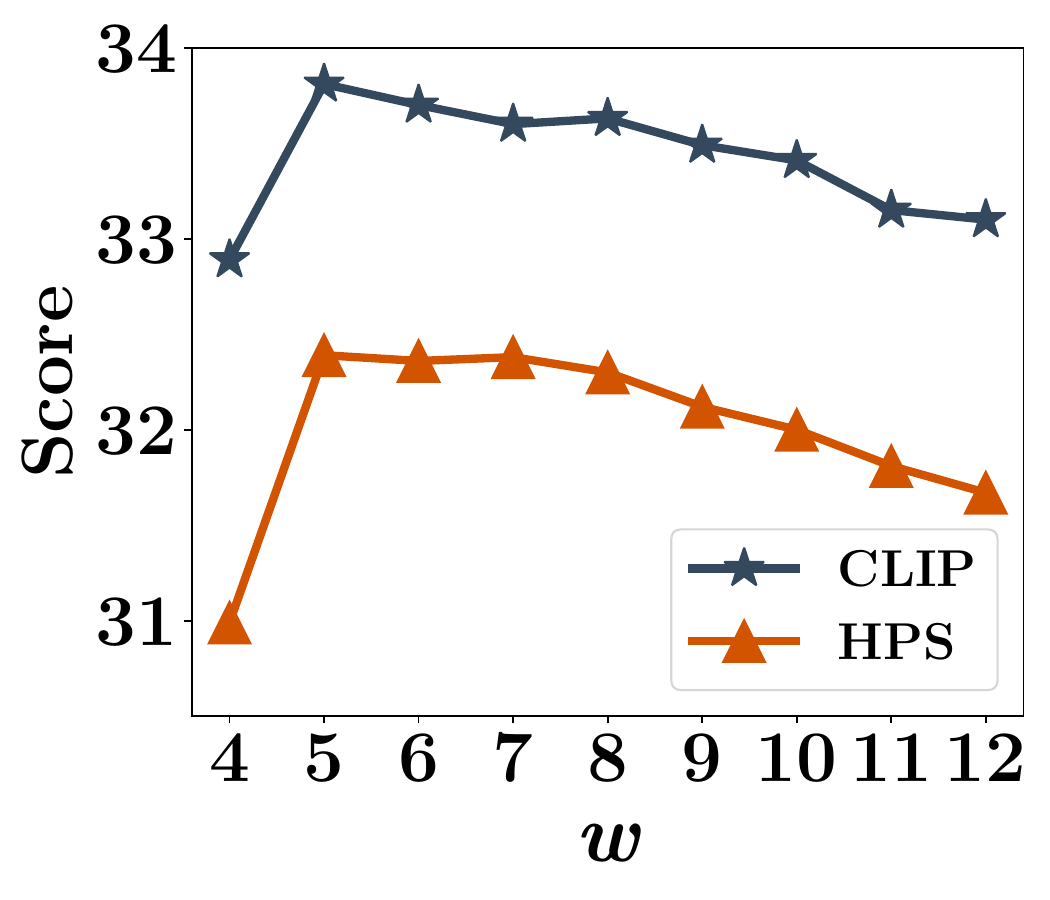}
        \caption{Guidance scale $w$}
        \label{fig:ablate_w_sub}
    \end{subfigure}
    \hfill
    \begin{subfigure}[b]{0.495\linewidth}
        \centering
        \includegraphics[width=\linewidth]{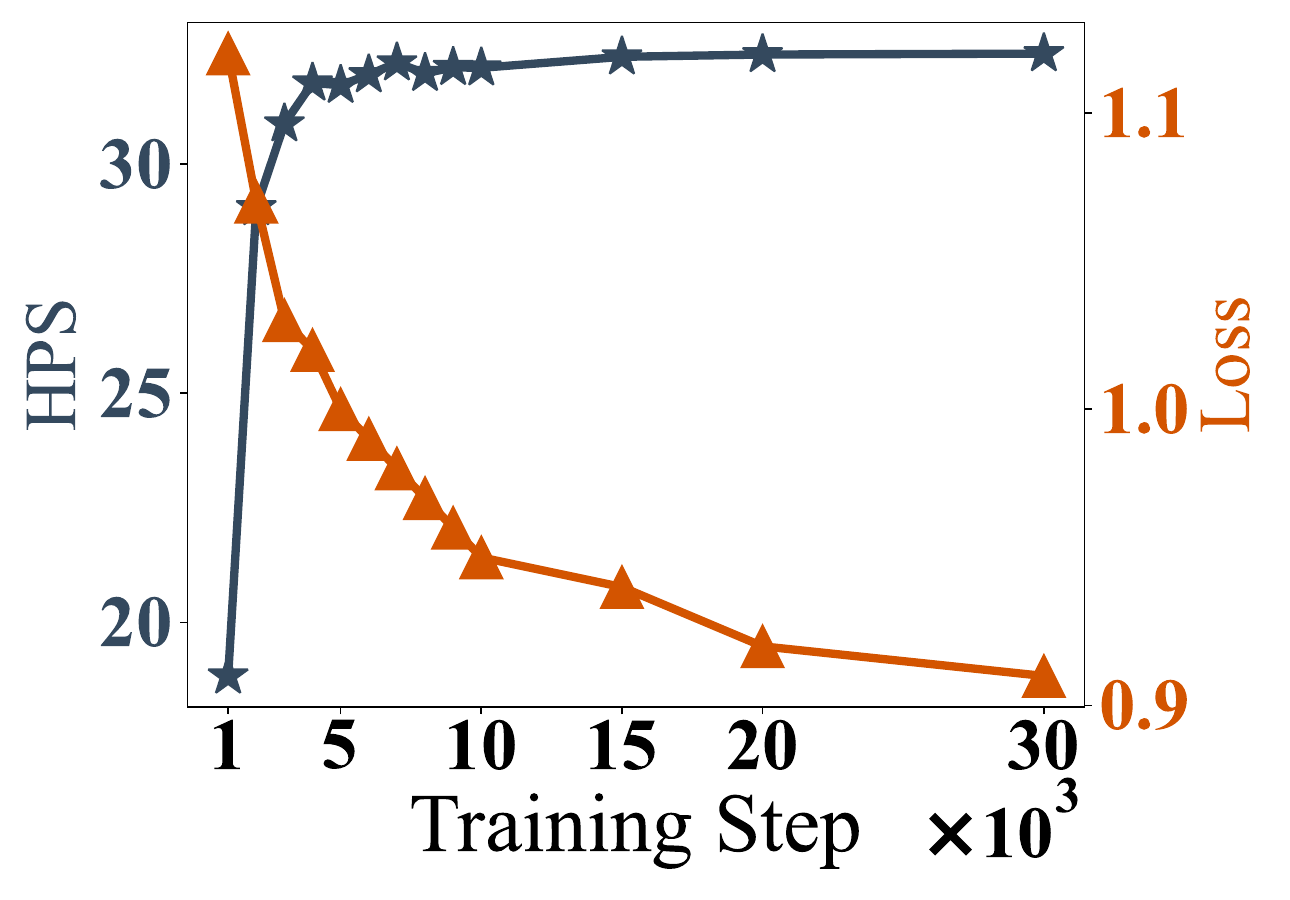}
        \caption{Training Step}
        \label{fig:ablate_loss_sub}
    \end{subfigure}
    \caption{Ablation study on varying (a) guidance scale $w$ and (b) the number of training steps in In-house T2I.}
    \label{fig:ablate_w}
\end{figure}

\textbf{Modular Ablation.} We conduct modular ablation on In-house T2I using prompts from the Anime (see Table~\ref{tab:modular_ablation}). By replacing the additional MLP used in DistillCFG with $\mathcal{G}(\psi(w))$, we observe an improvement in all scores. Furthermore, the complete TeEFusion configuration achieves the best results, consistently outperforming all other variants. These findings demonstrate that each component within TeEFusion meaningfully contributes to the effectiveness of guidance distillation. Notably, the student model employs simple Euler sampling without CFG. 

\textbf{Switching Teacher Sampling.} We also investigate the effectiveness of TeEFusion under different sampling strategies used by teacher models (cf. Table~\ref{tab:sampling_strategy_ablation}). The results indicate that TeEFusion achieves nearly lossless distillation regardless of the teacher sampling method. Although we observe minor performance reductions when distilling from simpler sampling methods in Euler+CFG and Z-Sampling+CFG, these slight differences are expected, as perfectly matching the teacher model is challenging. 

Importantly, when distilling from the more robust sampling method (W2SD+CFG), TeEFusion not only achieves near-lossless distillation but slightly surpasses the teacher’s performance across several metrics. This demonstrates that TeEFusion effectively preserves model quality during distillation. Moreover, these results indicate a promising property of TeEFusion: its performance is expected to further improve as stronger sampling strategies or larger teacher models become available in the future, making our method highly scalable and adaptable for test-time scaling.

\begin{figure}
    \centering
    \begin{subfigure}[b]{0.45\linewidth}
        \centering
        \includegraphics[width=\linewidth]{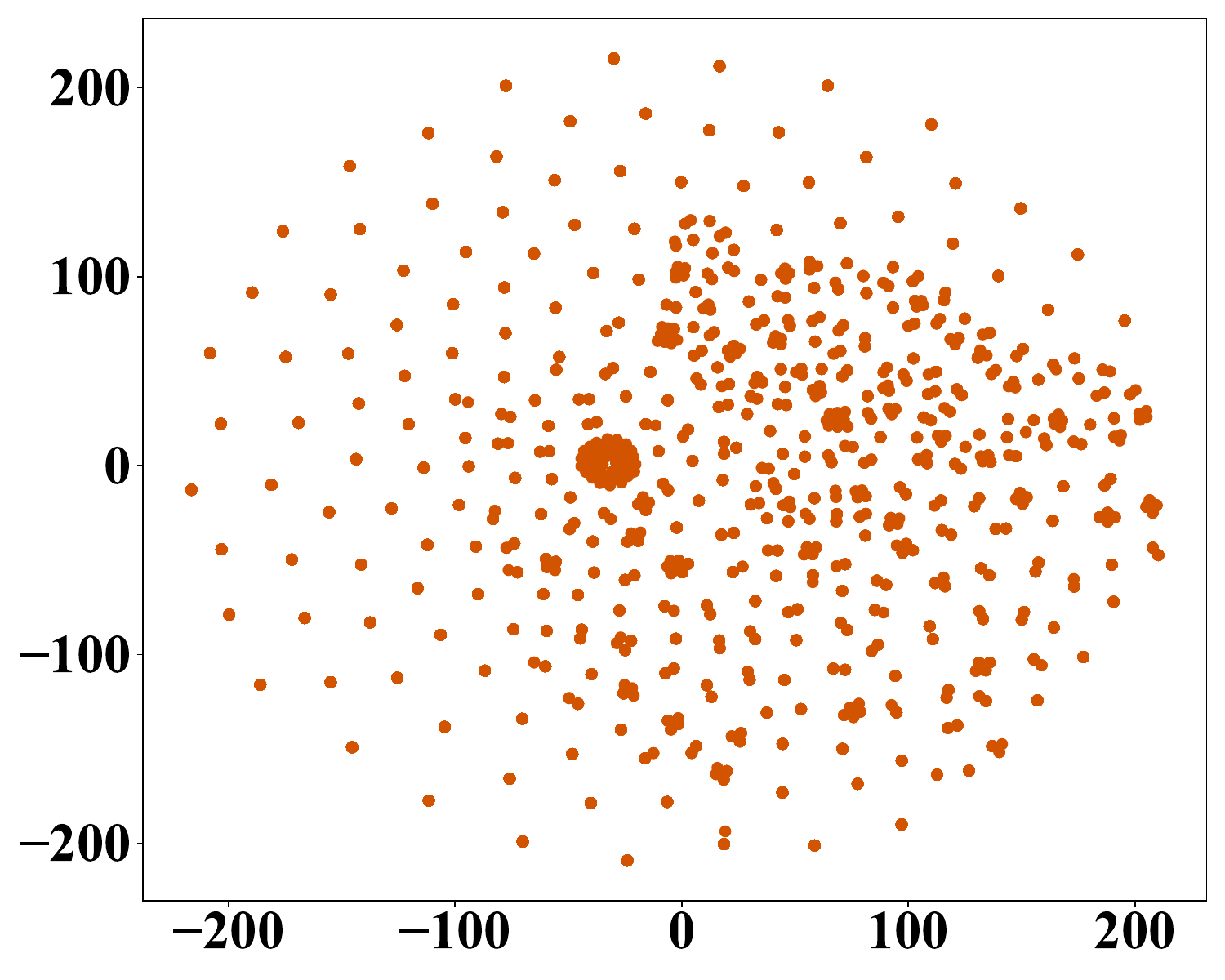}
        \caption{DistillCFG}
        \label{fig:tsne_distillcfg}
    \end{subfigure}
    \hfill
    \begin{subfigure}[b]{0.45\linewidth}
        \centering
        \includegraphics[width=\linewidth]{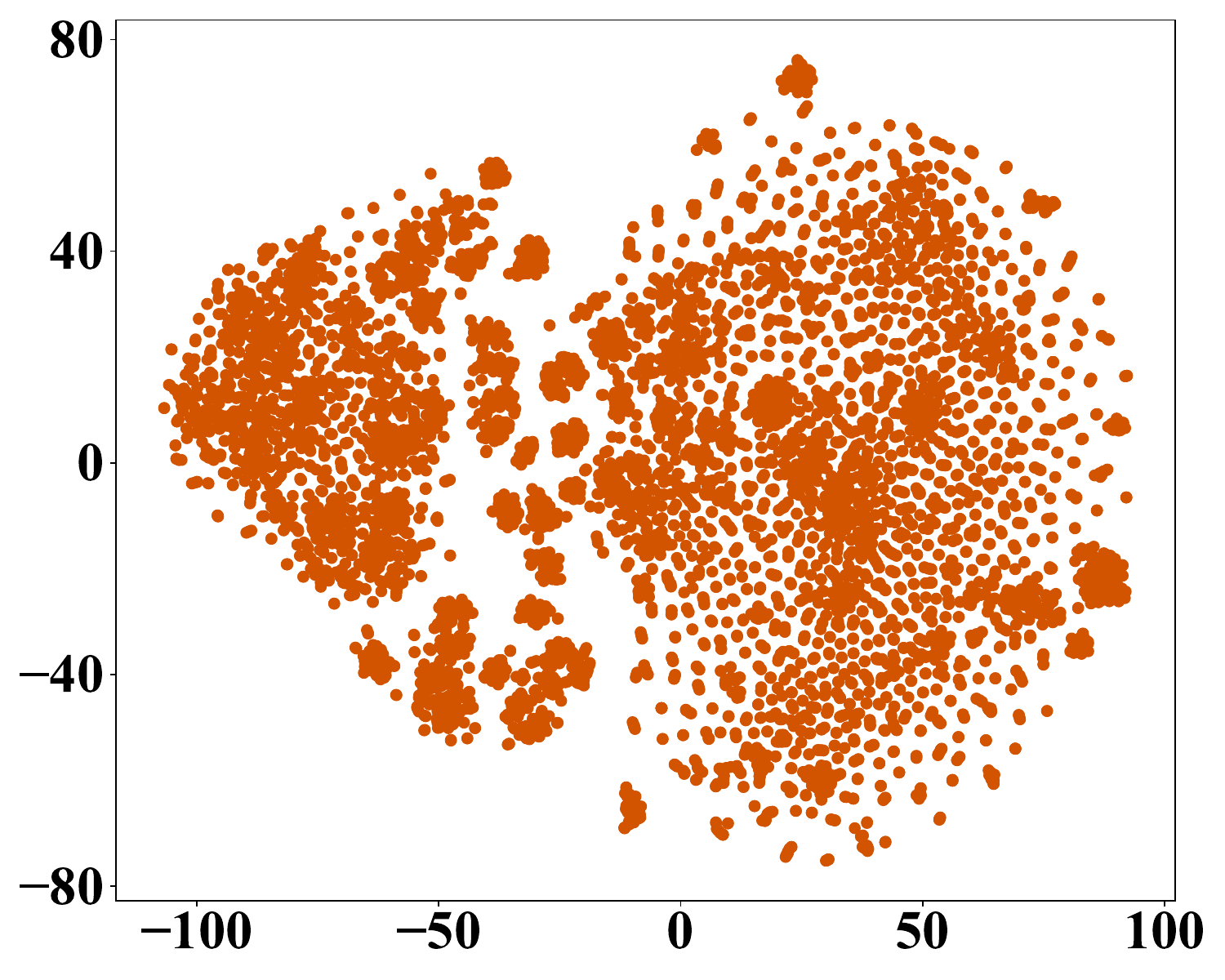}
        \caption{TeEFusion}
        \label{fig:tsne_teefuion}
    \end{subfigure}
    \caption{Illustration of embedded $w$, with values randomly sampled from the range [2, 14], visualized using t-SNE~\cite{tsne}.}
    \label{fig:tsne}
\end{figure}

\textbf{Is $w$-distillation successful?} We further examine the influence of varying the guidance scale $w$ on TeEFusion's performance. As illustrated in Fig.~\ref{fig:w_compare}, the images produced by TeEFusion under different $w$ settings are both diverse and visually appealing. Moreover, the model achieves the highest HPS and CLIP scores at $w=5$ (see Fig.~\ref{fig:ablate_w_sub}), indicating that an optimal guidance scale effectively enhances both aesthetic quality and prompt adherence. The distinct score variations across different $w$ values further confirm that TeEFusion maintains sensitivity to guidance scale changes after distillation. These findings validate the capability of our approach to preserve meaningful guidance scale characteristics, thereby enabling users to effectively balance text-image consistency and visual aesthetics.

\textbf{How fast does TeEFusion converge?} We further examine the convergence speed of TeEFusion in Fig.~\ref{fig:ablate_loss_sub}. It reveals that the HPS value increases rapidly during the early stages of training (below 5k steps) and then gradually saturates. Similarly, the loss curve drops quickly at first and levels off once the training steps exceed 10k. These findings clearly indicate that TeEFusion converges remarkably fast. This rapid convergence can be attributed to the fact that all encoding operations related to the guidance scale $w$ utilize modules directly inherited from a pretrained model, without any randomly initialized components. In fact, for In-house T2I, TeEFusion converges in under 4 hours when trained on 16 A100 GPUs, demonstrating an exceptionally efficient training process.

\textbf{Qualitative Analysis.} In Fig.~\ref{fig:w_compare}, we present the qualitative results of various methods across different $w$. We observe that the images generated by the teacher model using W2SD+CFG do not consistently achieve optimal aesthetic quality, particularly when $w$ is excessively high. Notably, at high $w$ values, W2SD+CFG is more prone to collapse. For example, the left image appears less appealing, while the right image exhibits pronounced periodic artifacts~\cite{artifacts}. In contrast, the distilled models do not suffer from these issues, which explains why TeEFusion attains higher aesthetic scores compared to W2SD+CFG due to its enhanced stability. Moreover, we consistently observed that TeEFusion outperforms DistillCFG, especially at lower $w$ values.

Additionally, in Fig.~\ref{fig:tsne} we illustrate the feature distributions of embedded $w$. An interesting observation is that the embedded $w$ obtained by TeEFusion exhibit a more coherent distribution, whereas those of DistillCFG appear markedly more discrete. This can be attributed to TeEFusion's additional utilization of information from the $c-\varnothing$, which results in a smoother guidance signal and, consequently, improved generation quality.

\section{Conclusions and Limitations}

In this paper, we introduced TeEFusion, an efficient distillation method that fuses guidance magnitudes into text embeddings, streamlining the inference process for text-to-image synthesis. By linearly combining conditional and unconditional embeddings, TeEFusion replicates the teacher model’s performance, achieving comparable image quality while enabling up to 6$\times$ faster inference. However, TeEFusion sometimes produces semantic inconsistencies (e.g., mismatched attributes) and its outputs do not always precisely align with the teacher model (cf.~Fig.~\ref{fig:w_compare}). Future work will focus on mitigating these issues.

{
    \small
    \bibliographystyle{ieeenat_fullname}
    \bibliography{main}
}

\clearpage
\appendix
\section{More Experimental Results and Analyses}

\subsection{Quantitative Analysis of Additive Text Embeddings}
\label{sec:cosine_similarity}
To validate the effectiveness of additive text embeddings, we conducted quantitative experiments across different text-to-image models. The cosine similarity between original and fused embeddings (Cos Sim.$_{\text{txt}}$) and their corresponding generated images (Cos Sim.$_{\text{img}}$) are summarized in the table below:

\begin{center}
\footnotesize
\begin{tabular}{l|ccc}
\hline
Metric & SD3 & In-house T2I & FLUX.1-dev \\
\hline
$\text{Cos Sim.}_{\text{txt}}$ & 0.8073 & 0.8192 & 0.8286 \\
$\text{Cos Sim.}_{\text{img}}$ & 0.8732 & 0.9137 & 0.9318 \\
\hline
\end{tabular}
\end{center}

These results confirm that additive embedding operations preserve over 80\% cosine similarity in text space and over 90\% in image space, demonstrating their ability to merge diverse semantic patterns effectively.

\subsection{Operational Boundaries and Failure Cases}
Our fusion mechanism $\mathcal{G}(\psi(w))\,\mathcal{F}(c-\varnothing)$ operates within the encoder's linear regime through bounded sine-cosine positional encodings ($\|\mathcal{G}(\psi(w))\,\mathcal{F}(c-\varnothing)\|_2 \leq \delta$). 
However, failure cases arise when:
\begin{itemize}
    \item Semantic vectors exhibit non-orthogonality (e.g., contradictory phrases like ``cold fire'')
    \item Contextual interference occurs in composite prompts (e.g., ``not a cat'')
\end{itemize}

\begin{figure}
    \centering
    \includegraphics[width=.95\linewidth]{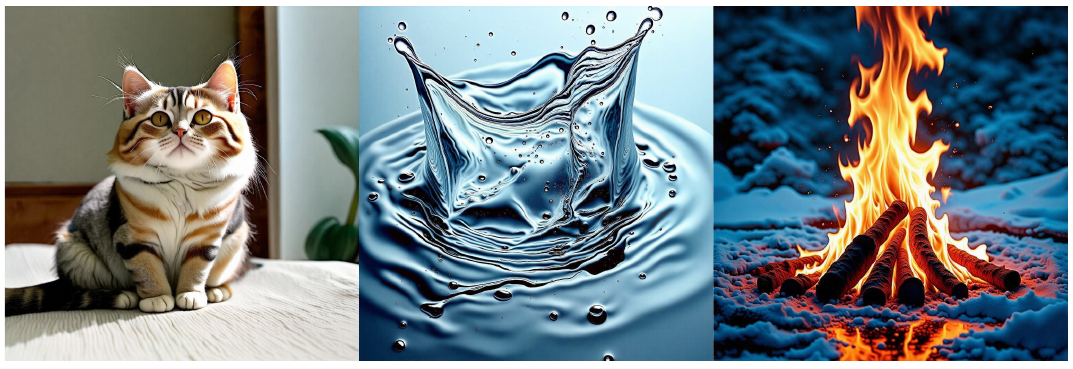}
    \caption{Generation examples of failure cases. Prompt: \textit{1) not a cat. 2) liquid glass. 3) cold fire.}}
    \label{fig:failure}
\end{figure}

These limitations are visualized in Figure~\ref{fig:failure}.

\end{document}